\documentclass[conference]{IEEEtran}
\IEEEoverridecommandlockouts
\usepackage{cite}
\usepackage{amsmath,amssymb,amsfonts}
\usepackage{algorithmic}
\usepackage{booktabs}
\usepackage{multirow}
\usepackage{graphicx}
\usepackage{textcomp}
\usepackage{comment}
\usepackage{lipsum}
\usepackage{threeparttable}
\usepackage{hyperref}
\usepackage{xcolor}
\usepackage{tabularx}

\def\BibTeX{{\rm B\kern-.05em{\sc i\kern-.025em b}\kern-.08em
    T\kern-.1667em\lower.7ex\hbox{E}\kern-.125emX}}
\begin{document}

\title{SSA-UNet: Advanced Precipitation Nowcasting via Channel Shuffling}

    \author{
    \IEEEauthorblockN{
        Marco Turzi, Siamak Mehrkanoon\IEEEauthorrefmark{1}\thanks{*Corresponding author.} 
    }
    \IEEEauthorblockA{
    Department of Information and Computing Sciences, Utrecht University, Utrecht, Netherlands}
    {m.turzi@students.uu.nl, 
    s.mehrkanoon@uu.nl}
}

\maketitle

\begin{abstract}
Weather forecasting is essential for facilitating diverse socio-economic activity and environmental conservation initiatives. Deep learning techniques are increasingly being explored as complementary approaches to Numerical Weather Prediction (NWP) models, offering potential benefits such as reduced complexity and enhanced adaptability in specific applications. This work presents a novel design, Small Shuffled Attention UNet (SSA-UNet), which enhances SmaAt-UNet's architecture by including a shuffle channeling mechanism to optimize performance and diminish complexity. To assess its efficacy, this architecture and its reduced variant are examined and trained on two datasets: a Dutch precipitation dataset from 2016 to 2019, and a French cloud cover dataset containing radar images from 2017 to 2018. Three output configurations of the proposed architecture are evaluated, yielding outputs of 1, 6, and 12 precipitation maps, respectively. To better understand how this model operates and produces its predictions, a gradient-based approach called Grad-CAM is used to analyze the outputs generated. The analysis of heatmaps generated by Grad-CAM facilitated the identification of regions within the input maps that the model considers most informative for generating its predictions. The implementation of SSA-UNet can be found on our Github\footnote{\href{https://github.com/MarcoTurzi/SSA-UNet}{https://github.com/MarcoTurzi/SSA-UNet}}. 
\end{abstract}


\section{Introduction}
Weather forecasting is an indispensable domain, deemed crucial for various operations, including aviation safety, emergency response, agricultural planning, maritime navigation, and outdoor event management, in addition to improving public safety. Furthermore, accurate weather forecasting can significantly help mitigate the pollution from heavy-vehicle traffic. The author in\cite{cools2010assessing} showed that severe weather can significantly increase vehicle utilization and traffic congestion. Consequently, accurate precipitation nowcasting could help people avoid superfluous vehicle journeys, thus alleviating traffic congestion and its related impacts on the environment.

Given the growing significance of precipitation nowcasting in several application domains, it is essential to improve the accessibility of weather forecasting models even when available resources are limited. Currently, Numerical Weather Prediction (NWP) models represent the main instruments for weather prediction. However, these models are computationally demanding and require considerable processing power and time, which constrains their accessibility and responsiveness, particularly for short-term predictions or nowcasting\cite{lahouar2017hour}.

As the field of artificial intelligence expands, numerous applications of deep learning have demonstrated remarkable performance across various domains. For example, deep learning has been used to forecast air quality\cite{li2016deep}, traffic flow\cite{lv2014traffic}, tidal level estimation\cite{riazi2020accurate}, and weather prediction\cite{baboo2010efficient,kamilaris2018deep,mehrkanoon2019deep,singh2019weather,vatamany2025graph}.

Initially, Recurrent Neural Networks (RNN) became widely recognized among researchers as they were specifically suited to deal with time series. However, the gradient vanishing issue and their inability to capture long-term dependencies forced academics to look for different approaches. Convolutional architectures gained popularity because of their efficacy in capturing spatial relationships in meteorological data, including patterns in satellite imagery and radar maps. Models such as UNet \cite{ronneberger2015u}, ResNet \cite{he2016deep}, and VGGNet \cite{simonyan2014very} have shown remarkable efficacy in object detection and image segmentation tasks. Moreover, ConvLSTM \cite{shi2015convolutional} has also shown great performance by integrating the advantages of recurrent and convolutional networks, which makes it particularly effective in tasks requiring understanding of spatial patterns and temporal dynamics.

Interpretability poses a significant challenge for AI, potentially delaying its deployment and acceptance in vital sectors such as healthcare, finance, and autonomous driving, where understanding the decision-making process is crucial to fostering confidence and assuring responsibility. Explainable artificial intelligence (XAI) approaches are crucial for fostering user trust and improving understanding of how these models produce their predictions. XAI seeks to create techniques to explain the predictions of AI models by highlighting significant aspects and providing visual representations of model behavior \cite{arrieta2020explainable}. Techniques like Class Activation Mapping (CAM) and Gradient-weighted Class Activation Mapping (Grad-CAM) are extensively utilized to produce heatmaps that highlight the areas of an input image that are most significant to a model's predictions, giving insight into their functioning.

This paper presents a novel architecture named SSA-UNet, which uses SmaAt-UNet as the core model and equips it with a channel and spatial attention mechanism known as Shuffle Attention\cite{zhang2021sa}, along with a shuffled variant of the depthwise separable convolution devised in this work. This variant partially substitutes the conventional depthwise separable convolutions in the encoder branch of SmaAt-UNet. This architecture exhibits enhanced performance compared to SmaAt-UNet, while decreasing the number of trainable parameters. Two configurations of this architecture are presented: the first realizes a 5\% reduction in parameters and outperforms SmaAt-UNet, while the second attains a significant 20\% decrease in parameters and demonstrates comparable performance to SmaAt-UNet. Both models are trained and evaluated using a precipitation map dataset from the Netherlands and a cloud cover dataset from France. Ultimately, Grad-CAM is utilized to explain the predictions of SSA-UNet. The heatmaps produced indicate the regions considered the most informative by the model and its layers, providing insight into the features that affect its predictions.

\section{Related Work}
The implementation of Deep Learning methods to perform weather forecasting tasks has become increasingly popular. Methods such as Recurrent Neural Network (RNN)\cite{rumelhart1986learning} and Long Short-Term Memory (LSTM)\cite{hochreiter1997long} have been widely employed in weather forecasting tasks due to their innate ability to deal with time-series datasets. In \cite{xu2019satellite}, a Generative Adversarial Network (GAN) is trained to generate satellite cloud images from random data. The GAN’s generator is then used to produce future cloud images by taking inputs from an LSTM model, which itself processes previous cloud images.

The use of convolution in weather forecasting tasks has become widespread because of its ability to effectively capture spatial patterns and local features. In ConvLSTM \cite{shi2015convolutional}, the fully connected layers within a traditional LSTM are replaced with convolutional layers, enabling it to perform precipitation nowcasting more effectively. This structure has been shown to outperform classic LSTMs in capturing spatio-temporal patterns in weather data. In \cite{mehrkanoon2019deep}, deep convolutional neural network models are used for temperature and wind speed forecasting, leveraging spatio-temporal multivariate weather data. The models, which include 1D-, 2D-, and 3D-CNN architectures, learn shared representations from historical data and demonstrate improved prediction accuracy over classical neural network approaches. The authors of\cite{trebing2020wind} proposed a model that applies depthwise separable convolution on each of the three dimensions of the input data to perform wind speed prediction, demonstrating the ability of this model to learn better data representations. 
 
 In \cite{ronneberger2015u}, a novel architecture called UNet was introduced, which demonstrated impressive performance in image segmentation tasks. Trebing et al. \cite{trebing2021smaat} revisited the UNet architecture for precipitation nowcasting, managing to considerably reduce the number of parameters while maintaining comparable performances. To achieve this, they replaced all convolutions with a depthwise separable convolution \cite{chollet2017xception}, which uses grouped convolution to apply a single convolutional filter per input channel, and added a Convolutional Block Attention Module (CBAM) \cite{woo2018cbam}, which sequentially applies channel and spatial attention mechanisms to enhance feature representation. The authors of \cite{yang2022aa} adopted a similar approach, enhancing the TransUNet\cite{chen2021transunet} architecture with CBAM and depthwise-separable convolutions, achieving remarkable results in nowcasting tasks. In \cite{kaparakis2023wf} a two-stream UNet is proposed, where each stream separately processes precipitation and wind speed radar images. The outputs are then combined to predict future precipitation maps. Fernández et al. \cite{fernandez2021broad} enhanced UNet with Atrous Spatial Pyramid Pooling, enabling the model to learn multi-scale features.  

A common challenge with all the methodologies discussed above is their inherent tendency to lack transparency. Their complexity can make understanding their functioning quite challenging. Many scholars have been trying to make these models more transparent and interpretable to increase user trust. However, explaining the predictions that these models generate goes beyond simply increasing user confidence. For instance, by implementing a transparent machine learning approach based on fundamental principles of ocean physics, Sonnewald et al. \cite{sonnewald2021revealing} gained insight into the effects of global warming on the weakening of North Atlantic ocean circulation. 

Many methods have been devised to achieve explainability. One of the most well-known methods is Local Interpretable Model-agnostic Explanation (LIME) \cite{ribeiro2016should}, which provides local interpretability by approximating complex models with simpler, interpretable models in the vicinity of a given prediction. In \cite{gibson2021training}, LIME was used to explain each individual seasonal precipitation forecast and identify the factors that led the model to make incorrect predictions. Gradient-based approaches such as Gradient-weighted Class-Activation Mapping (Grad-CAM) \cite{selvaraju2017grad} and Integrated Gradients (IG) \cite{sundararajan2017axiomatic} have also been extensively used to explain the predictions of weather forecasting models. These methods determine the input areas the model uses to generate a prediction by analyzing the output gradients with respect to the input features. In \cite{renault2023sar} \cite{reulen2024ga}, the authors sought to explain the precipitation maps produced by their UNet structures by analyzing the heatmaps generated with Grad-CAM.

\begin{figure*}[h!]
    \centering
    \includegraphics[width=\textwidth,height=0.4995\linewidth]{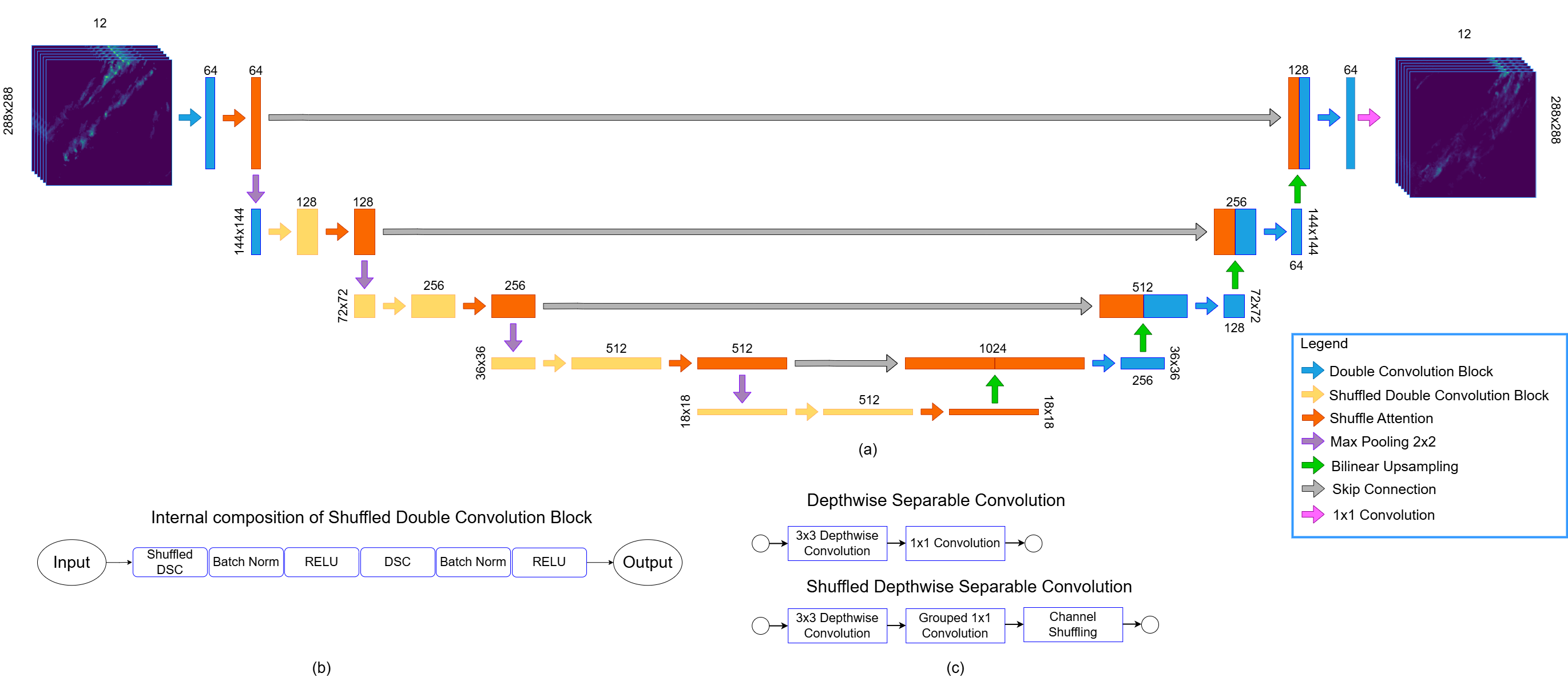} 
    \caption{(a) Visual representation of SSA-UNet. The rectangles indicate the data dimensions, with height representing the spatial dimension and width representing the channel dimension. Arrows and their color specify the type of operation applied to the input data. (b) Shows the composition of the shuffled double convolution block. In this block, the input passes through a sequence consisting of a shuffled depthwise separable convolution, followed by batch normalization, ReLU activation, a depthwise separable convolution, and another round of batch normalization and ReLU activation. (c) Illustrates the sequence of operations that are performed by the depthwise separable convolution and the shuffled depthwise separable convolution.}
    \label{fig:fullwidth}
\end{figure*}

\section{Method}
\subsection{Proposed SSA-UNet Model}
\subsubsection{Architecture}
     As shown in Fig. \ref{fig:fullwidth}(a), SSA-UNet consists of an encoder and a decoder branch that perform complementary operations on input images. The encoder captures high-level spatial features by reducing spatial resolution and increasing channel depth, while the decoder reconstructs the image by combining features (channels) from the encoder and restoring spatial resolution. The encoder branch is structured into five levels, while the decoder one into 4. Each encoder level is composed of a shuffled double convolution block (indicated by the yellow arrow), except for the first layer, which employs a classic double convolution block (blue arrow). Furthermore, each level includes the Shuffle Attention module (orange arrow) and a 2×2 max-pooling layer (purple arrow). The shuffled double convolution block increases the number of channels, and the Shuffle 
 Attention module highlights the most informative regions of the image. The output size is then halved by the max-pooling layer and transferred to the next encoder level. Concurrently, the same output is also passed to the corresponding decoder level via skip connections (grey arrow).
The decoder consists of a bilinear upsampling module (green arrow) that takes as input the concatenation of the previous level's feature map and the output of the Shuffle Attention module from the corresponding encoder level, doubling the image size. This is followed by a double convolution block that reduces the number of channels by half. Finally, a 1x1 convolution (pink arrow) is implemented in the decoder to output the feature maps that represent the model predictions.

\subsubsection{Shuffled Double Convolution Block}
\begin{figure}[h!]
    \centering
    \includegraphics[width=0.6\linewidth]{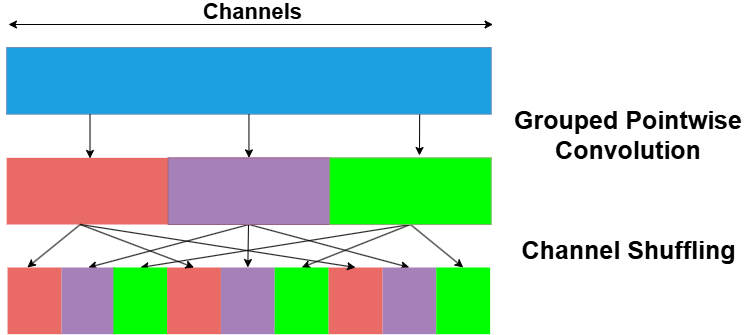} 
    \caption{Illustration of the impact of a grouped pointwise convolution followed by channel shuffling on the input channels. The grouped convolution processes channel groups independently, while channel shuffling redistributes channels to enable cross-group information exchange.}
    \label{shuffled}
\end{figure}

As shown in the previous section, this architecture uses two versions of the double convolution block: the one proposed by \cite{trebing2021smaat} and a novel variant introduced in this work. The new variant incorporates channel shuffling to reduce the number of trainable parameters. Fig. \ref{fig:fullwidth}(c) offers a visualization of the two depthwise separable convolution modules. The primary difference between the two blocks is that the shuffled variant replaces the first depthwise separable convolution with a shuffled depthwise separable convolution. Specifically, this involves introducing a grouped pointwise convolution after the 3x3 depthwise convolution, followed by channel shuffling. Throughout this process this module manages to use less parameters and redistributes the channels yielded by every group. This mechanism enhances feature interactions across channel groups by ensuring that the next layer can utilize channels derived from a broader range of the input, rather than being limited to a small subset of it. Specifically, the outputs of the different groups are divided uniformly and redistributed across the entire channel dimension, as illustrated in Fig. \ref{shuffled}. The shuffled double convolution block is used in the encoder branch from the second to the fifth levels. In these blocks, the group size for the grouped pointwise convolutions is set to 16 for the second and third levels, and 32 for the fourth and fifth levels. Its implementation in the first level was not feasible due to an insufficient number of input channels for effective group convolution. Similarly, it was excluded from the decoder branch to prevent an excessive reduction in parameters, which could have resulted in the model learning a less effective representation of the data. A visualization of the entire composition of the module is shown in Fig. \ref{fig:fullwidth}(b).

\subsubsection{Shuffle Attention Module}
To enhance the identification of informative regions within images, the Shuffle Attention module is integrated after each shuffled double convolution block in the encoder. At first glance, the Shuffle Attention module may seem similar to CBAM, which is used in SmaAt-UNet\cite{trebing2021smaat}, since they both perform channel and spatial attention. However, their mechanisms differ significantly. CBAM applies channel attention and spatial attention sequentially, whereas the Shuffle Attention module processes them in parallel. Specifically, it splits the input into two parts: one for channel attention and the other for spatial attention. The outputs are then merged, and the channels are shuffled to enable cross-group information flow. Additionally, the Shuffle Attention module is highly efficient, requiring only a small fraction of the trainable parameters used by CBAM, making it a lightweight and effective block.

\subsubsection{Model Variation}
To evaluate the effectiveness of this architecture, two versions are trained and tested: SSA-UNet and a reduced variant called SSA-UNet(reduced). These models have the same overall structure and settings but differ in the number of kernels per layer in the shuffled depthwise separable convolutions. Although this adjustment may seem minor, it significantly impacts the number of trainable parameters. Specifically, SSA-UNet uses 3 kernels per layer, resulting in 3.8 million parameters, whereas SSA-UNet(reduced) employs 2 kernels per layer, reducing the parameter count to 3.1 million.


\subsection{Training}Following the lines of \cite{trebing2021smaat}, the dataset is divided into small batches of 6 maps each and the model is trained for up to 200 epochs. An early stopping mechanism stops the training process when the validation loss does not show improvement over a span of 15 consecutive epochs. The Adam optimizer is used with an initial learning rate of 0.001, which is reduced by a factor of 0.1 if the validation loss does not improve for 4 consecutive epochs. The model is trained on an Nvidia A10 GPU.

\subsection{Evaluation}
Evaluating the performance of these models in their different configurations primarily involves using the Mean Squared Error (MSE), as it is an appropriate metric for regression tasks such as the one studied here. Furthermore, following the lines of \cite{trebing2021smaat}, each pixel of the output images is binarized to count the True Positives, False Positives, True Negatives, and False Negatives. This enables the calculation of additional important metrics, including precision, accuracy, recall, and F1-score. The performance of these models is further evaluated in comparison with SmaAt-UNet and the Persistence method, the latter of which uses only the final input image as the predicted outcome.

\subsection{Explainability}
Deep neural networks are often criticized for their lack of transparency, which makes difficult to understand how they produce certain predictions. Grad-CAM (Gradient-weighted Class Activation Mapping) is a powerful gradient-based explainability technique that addresses this issue by identifying and visualizing the regions of an input image that most strongly influence the network's prediction. The importance weights for neurons are determined by calculating the gradient of the class score with respect to feature map activations. Subsequently, a linear combination of these activations is followed by ReLU to derive the final output. This produces a heatmap that visually emphasizes the parts of the image that significantly influence the prediction, thus providing a transparent explanation. The heatmaps derived from the gradients produced by the 5 levels of the encoder and the 4 levels of the decoder were specifically analyzed in our experiments. The analysis was conducted on the shuffled double convolution block, the shuffled and classic depthwise separable convolutions, and the Shuffle Attention module within the encoder, while only on the double convolution block in the decoder.

\begin{table}[htbp]
    \centering
    \caption{Comparison of  Parameters, and Inference Time for SMaAt-UNet, SSA-UNet, and SSA-UNet(reduced).}
    \begin{tabular}{|l|c|c|}
        \hline
        \textbf{Model}  & \textbf{Parameters} & \textbf{Inference Time (ms)} \\
        \hline
        SmaAt-UNet & 4M & 50.00 \\
        SSA-UNet & 3.8M & 47.48 \\
        SSA-UNet(reduced*) & 3.1M & 44.72 \\
        \hline
    \end{tabular}
    \label{params}
    \begin{tablenotes}
     \item $\star$ reduced version using 2 kernels per layer instead of 3.
\end{tablenotes}
\end{table}

\begin{table*}[ht!]
\centering
\scriptsize
\caption{Performance comparison of Persistence method, SmaAt-UNet, SSA-UNet and SSA-UNet(reduced) for the NL-50 and NL-20 test datasets across different output configurations. Metrics include MSE (pixel), precision, recall, accuracy, and F1 score. Best values for each metric are highlighted in \textbf{bold}.}
\begin{tabularx}{\textwidth}{>{\centering\arraybackslash}p{1.5cm} >{\centering\arraybackslash}p{2cm} >{\centering\arraybackslash}p{2.5cm} >{\centering\arraybackslash}p{1.5cm} >{\centering\arraybackslash}p{1.5cm} >{\centering\arraybackslash}p{1.5cm} >{\centering\arraybackslash}p{1.5cm} >{\centering\arraybackslash}p{1.5cm}}
\toprule
\textbf{Dataset} & \textbf{Outputs} & \textbf{Model} & \textbf{MSE (pixel)} & \textbf{Precision} & \textbf{Recall} & \textbf{Accuracy} & \textbf{F1 Score} \\
\midrule
\multirow{12}{*}{NL-50} 
 & \multirow{4}{*}{\shortstack{12\\(5 to 60 min)}} & Persistence   & 0.0262 & 0.693 & 0.633 & 0.774 & 0.661 \\
 &                           & SmaAt-UNet   & 0.0132 & 0.637 & \textbf{0.903} & 0.787 & 0.747 \\
 &                           & SSA-UNet     & \textbf{0.0131} & 0.690 & 0.859 & 0.816 & 0.765 \\
 &                           & SSA-UNet(reduced*)   & 0.0133 & \textbf{0.713} & 0.847 & \textbf{0.828} & \textbf{0.774} \\
 
\cmidrule{2-8}
 & \multirow{4}{*}{\shortstack{6\\(5 to 30 min)}} & Persistence   & 0.0195 & 0.752 & 0.726 & 0.814 & 0.739 \\
 &                           & SmaAt-UNet   & 0.0117 & 0.756 & \textbf{0.903} & 0.860 & 0.823 \\
 &                           & SSA-UNet     & \textbf{0.0089} & \textbf{0.806} & 0.879 & \textbf{0.880} & \textbf{0.841} \\
 &                           & SSA-UNet(reduced*)   & 0.0098 & 0.774 & 0.892 & 0.867 & 0.828 \\
          
\cmidrule{2-8}
 & \multirow{4}{*}{\shortstack{1\\(30 min)}} & Persistence   & 0.0249 & 0.678 & 0.643 & 0.756 & 0.660 \\
 &                           & SmaAt-UNet   & 0.0126 & \textbf{0.741} & 0.824 & 0.829 & 0.780 \\
 &                           & SSA-UNet     & \textbf{0.0121} & \textbf{0.741} & 0.841 & \textbf{0.833} & 0.788 \\
 &                           & SSA-UNet(reduced*)   & \textbf{0.0121} & 0.730 & \textbf{0.863} & 0.832 & \textbf{0.791} \\
\midrule
\multirow{12}{*}{NL-20} 
 & \multirow{4}{*}{\shortstack{12\\(5 to 60 min)}} & Persistence   & 0.0207 & 0.579 & 0.551 & 0.838 & 0.565 \\
 &                           & SmaAt-UNet   & \textbf{0.0105} & 0.532 & \textbf{0.857} & 0.829 & 0.656 \\
 &                           & SSA-UNet     & \textbf{0.0105} & 0.602 & 0.794 & 0.861 & 0.685 \\
 &                           & SSA-UNet(reduced*)   & 0.0106 & \textbf{0.618} & 0.794 & \textbf{0.868} & \textbf{0.696} \\
\cmidrule{2-8}
 & \multirow{4}{*}{\shortstack{6\\(5 to 30 min)}} & Persistence   & 0.0185 & 0.658 & 0.646 & 0.866 & 0.652 \\
 &                           & SmaAt-UNet   & 0.0097 & 0.680 & \textbf{0.860} & 0.894 & 0.759 \\
 &                           & SSA-UNet     & \textbf{0.0085} & \textbf{0.739} & 0.831 & \textbf{0.910} & \textbf{0.782} \\
 &                           & SSA-UNet(reduced*)   & 0.0090 & 0.700 & 0.845 & 0.900 & 0.765 \\

\cmidrule{2-8}
 & \multirow{4}{*}{\shortstack{1\\(30 min)}} & Persistence   & 0.0228 & 0.559 & 0.543 & 0.827 & 0.551 \\
 &                           & SmaAt-UNet   & 0.0114 & \textbf{0.649} & 0.761 & \textbf{0.873} & 0.701 \\
 &                           & SSA-UNet     & \textbf{0.0112} & 0.597 & 0.789 & 0.855 & 0.680 \\
 &                           & SSA-UNet(reduced*)   & \textbf{0.0112} & 0.639 & \textbf{0.810} & \textbf{0.873} & \textbf{0.715} \\
 \bottomrule
\end{tabularx}
\label{merged_results_adjusted}
\begin{tablenotes}
     \item $\star$ reduced version using 2 kernels per layer instead of 3.
\end{tablenotes}
\end{table*}

\begin{table}[ht!]
\centering
\scriptsize
\caption{Comparison of Persistence method, SmaAt-UNet, SSA-UNet and SSA-UNet(reduced*) on various metrics for the cloud cover test dataset.}
\resizebox{\linewidth}{!}{
\begin{tabular}{lccccc}
\toprule
\textbf{Model} & \textbf{MSE (Pixel)} & \textbf{Precision} & \textbf{Recall} & \textbf{Accuracy} & \textbf{F1 Score} \\
\midrule
Persistence     & 0.1491               & 0.8722             & 0.8720          & 0.8509            & 0.8721 \\
SmaAt-UNet     & 0.0788               & \textbf{0.8998}             & 0.9122          & 0.8897            & 0.9060 \\
SSA-UNet       & \textbf{0.0785}               & 0.8948             & 0.9197          & \textbf{0.8901}            & 0.9071 \\
SSA-UNet(reduced*)     & 0.0788               & 0.8921             & \textbf{0.9228}          & 0.8899            & 0.9072 \\
\bottomrule
\end{tabular}
}
\label{cloud_results}
\begin{tablenotes}
     \item $\star$ reduced version using 2 kernels per layer instead of 3.
\end{tablenotes}
\end{table}

\section{Experiments}
\subsection{Precipitation dataset}
Similar to the approach in \cite{trebing2021smaat}, precipitation data from the Royal Netherlands Meteorological Institute (KNMI) is used, including nearly 420,000 rain maps collected at 5-minute intervals from 2016 to 2019. This dataset covers the Netherlands and adjacent areas and is produced by two C-band Doppler radar stations positioned in De Bilt and Den Helder.

The dataset is preprocessed in accordance with the lines of \cite{trebing2021smaat}. The raw rain maps (originally 765 × 700 pixels) are normalized by dividing by the highest value in the training set. Additionally, the maps are cropped to remove no-data pixels from the radar's maximum range, focusing on a central circular area of 421 × 421 pixels. Further center-cropping to 288 × 288 pixels is applied to eliminate remaining no-data areas, optimizing the input for model training. Two subsets are created: NL-50, where target images have at least 50\% rainy pixels, and NL-20, with at least 20\% rainy pixels.




Three different output configurations of our architecture are trained and evaluated. The first configuration predicts a single precipitation map 30 minutes ahead. The second generates six maps, predicting precipitation in 5-minute intervals up to 30 minutes ahead. The third configuration produces twelve maps, predicting precipitation in 5-minute intervals up to 60 minutes ahead. In all cases, the input consisted of 12 maps, representing the preceding hour of precipitation maps and stacked along the channel dimension. For each configuration, the models are trained on NL-50 and evaluated on both NL-50 and NL-20. The latter is used as an additional performance indicator to evaluate generalizability, as suggested in \cite{trebing2021smaat}. Checkpoints are saved after every epoch during training and selected those with lower validation loss for testing.

\subsection{Cloud Cover Dataset}
An additional evaluation of our model is conducted using the French cloud cover dataset mentioned in \cite{trebing2021smaat}. The dataset includes binary images with a resolution of 256x256 pixels, with each pixel assigned a value of 1 to indicate cloud presence and 0 to indicate absence. The images are recorded at 15-minute intervals. Four past images, indicative of the previous hour, are utilized as input to predict six future images, which represent the following 1.5 hours.

\begin{figure}[h!]
    \centering
    \includegraphics[width=\linewidth]{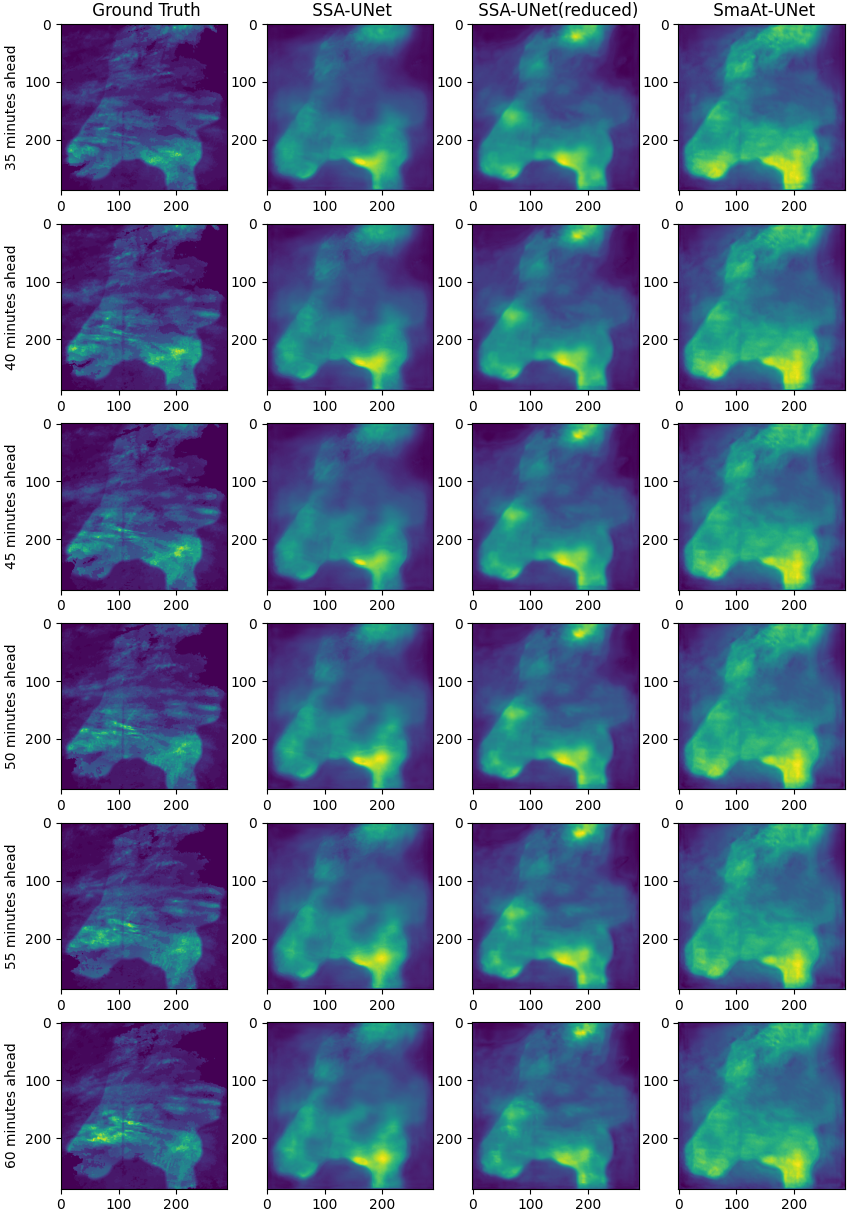}
    \caption{Comparison of predictions generated by SmaAt-UNet, SSA-UNet and SSA-UNet(reduced) with the ground truth. The predictions demonstrate the differences in performance between the architectures, where the SSA-UNet model shows better alignment with the true image.}
    \label{fig:predictions}
\end{figure}

\begin{figure}[h!]
    \centering
    \includegraphics[width=0.6\linewidth]{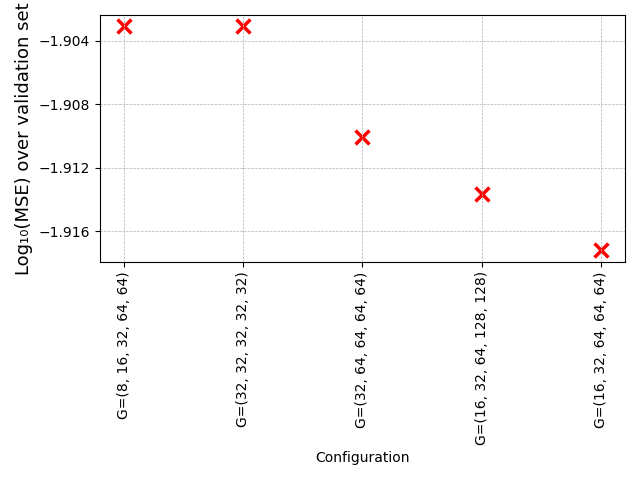}
    \caption{Visualization of the hyperparameter tuning results for SSA-UNet, showing the MSE values (in logarithmic scale) achieved by each tested configuration. Each configuration is defined as a list (G) of Shuffle Attention's group size values corresponding to each encoder level.}
    \label{fine_tuning}
\end{figure}

\section{Results and Discussion}

\subsection{Precipitation and Cloud cover results}

\subsubsection{Tuning}
Extensive hyperparameter tuning is conducted on the group size parameter of the Shuffle Attention mechanism at each level of SSA-UNet. Fig. \ref{fine_tuning} illustrates how the MSE loss fluctuates with different parameter settings. It is noticeable that the two configurations with the highest loss values are those with the smallest group size parameters overall. Analyzing the configurations whose parameter has been left constant in every level, they achieved the second and third worst score, suggesting that this kind of configuration does not allow the attention method to fully capture the local and spatial dependencies. Finally, the hybrid approach outperforms all other configurations, achieving slightly better results than the second incremental configuration.

\begin{figure}[ht!]
    \centering
    \includegraphics[width=\linewidth]{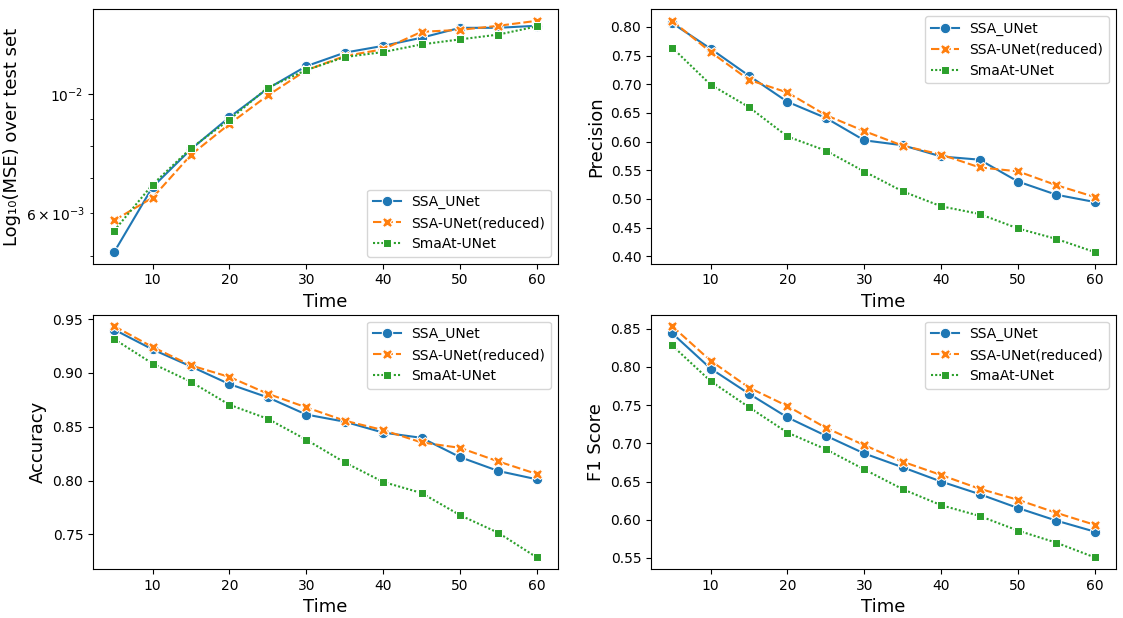}
    \caption{Performance metrics over time for SmaAt-UNet, SSA-UNet and SSA-UNet(reduced) in the 12-output configuration. The figure shows four plots representing the evolution of the MSE (in logarithmic scale), Accuracy, Precision, and F1 Score at 5-minute intervals over a 60-minute period. }
    \label{plot20}
\end{figure}

\begin{figure}[ht!]
    \centering
    \includegraphics[width=\linewidth]{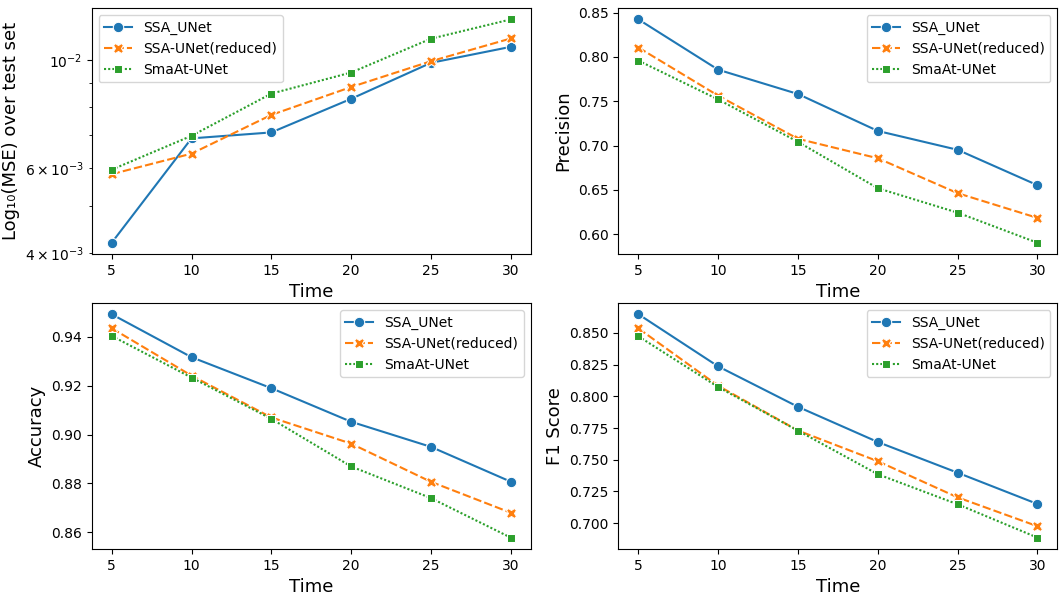}
    \caption{Performance metrics over time for SmaAt-UNet, SSA-UNet and SSA-UNet(reduced) in the 6-output configuration. The figure shows four plots representing the evolution of the MSE (in logarithmic scale), Accuracy, Precision, and F1 Score at 5-minute intervals over a 30-minute period. }
    \label{plot6}
\end{figure}

\subsubsection{Evaluation}
The experimental results on the NL-50 and NL-20 test datasets provide a clear demonstration of the advantages that channel shuffling brings to both the attention module and the depthwise separable convolution. The results are presented in Table \ref{merged_results_adjusted}. Focusing first on MSE, the primary metric used to evaluate our models, it is evident that both configurations consistently perform as well as or better than SmaAt-UNet. In particular, the only configuration where SSA-UNet significantly outperforms SmaAt-UNet is the 6-output configuration, where it achieves MSE values that are 25\% and 15\% lower than SmaAt-UNet in the NL-50 and NL-20 datasets, respectively. In this same output configuration, SSA-UNet(reduced) also succeeds in outperforming SmaAt-UNet, although with less significant margins compared to the improvements demonstrated by SSA-UNet. Although the differences in other configurations are less pronounced, the model's ability to maintain strong performance with fewer parameters is impressive.

In fact, as shown in Table \ref{params}, SSA-UNet and SSA-UNet(reduced) employ 3.8 million and 3.2 million trainable parameters, respectively. However, the number of parameters alone does not fully capture the efficiency of a model. As a result, the inference time is also analyzed, specifically the average time each model takes to generate predictions from a batch of six precipitation maps. This analysis confirms that the reduction in trainable parameters positively affects the model's speed.

Looking at Table \ref{merged_results_adjusted}, it may be hard to understand why increasing the number of output images from 1 to 6 improves performance, while increasing from 6 to 12 results in the opposite. However, it is important to consider that the initial precipitation maps are likely to closely resemble the last input map, making them easier to be predicted accurately. This naturally results in higher performance for those specific output maps, which increases the overall average performance across all output maps.  

It is important to note that, unlike the results from the 1-output configuration, the results for the other configurations represent the metrics averaged across all the output maps. For a more detailed representation of these configurations, the metrics are computed for each individual temporal step within the output. Fig. \ref{plot20} and Fig. \ref{plot6} present the variations of Mean Squared Error (MSE), Accuracy, Precision, and F1 Score over time for the 12-output and 6-output configurations, respectively. It is a common trend across all models for performance to decline as predictions extend further into the future. This is because maps farther in time compared to the input maps typically exhibit greater differences, making them more difficult to be predicted accurately. We can observe that for Accuracy, Precision, and F1 Score, SSA-UNet and SSA-UNet(reduced) show better results at each time step compared to SmaAt-UNet. In contrast, SmaAt-UNet exhibits better MSE values between 30 and 60 minutes ahead in the 12-output configuration. However, it is still outperformed by the other models in the 5 to 30-minute time frame in the 6-output configuration, where its performance is similar to that of the other models in the 12-output configuration.

Regarding the other metrics, SSA-UNet and SSA-UNet(reduced) outperform SmaAt-UNet in nearly all cases. Exceptions include the precision of the 1-output configuration, where SmaAt-UNet scores higher on the NL-20 dataset and matches our models on the NL-50 dataset. Additionally, it is worth noting that SmaAt-UNet consistently achieves a higher recall score. In addition to these observations, both SSA-UNet and SSA-UNet(reduced) exhibit significant superiority over the Persistence model across all configurations. This is noteworthy because in nowcasting tasks, where there are small temporal changes between the input and target frames, it can be particularly challenging to outperform this baseline \cite{soman2010review}. Our models consistently achieve around 50\% lower MSE than Persistence.

Fig. \ref{fig:predictions} presents the predictions generated by SSA-UNet, SSA-UNet(reduced), and SmaAt-UNet for forecasts ranging from 35 to 60 minutes ahead. It is evident that SSA-UNet produces precipitation maps that more closely resemble the ground truth. While all three architectures demonstrate sufficient accuracy in predicting the overall shape of precipitation areas, SSA-UNet outperforms the others in accurately identifying regions with higher precipitation intensity.

Similarly to the precipitation datasets, the proposed models outperform both SmaAt-UNet and Persistence on every metric except precision, where SmaAt-UNet scores better, in the cloud cover dataset. The results are displayed in Table  \ref{cloud_results}.

\begin{figure}[ht!]
    \centering
    \includegraphics[width=\linewidth]{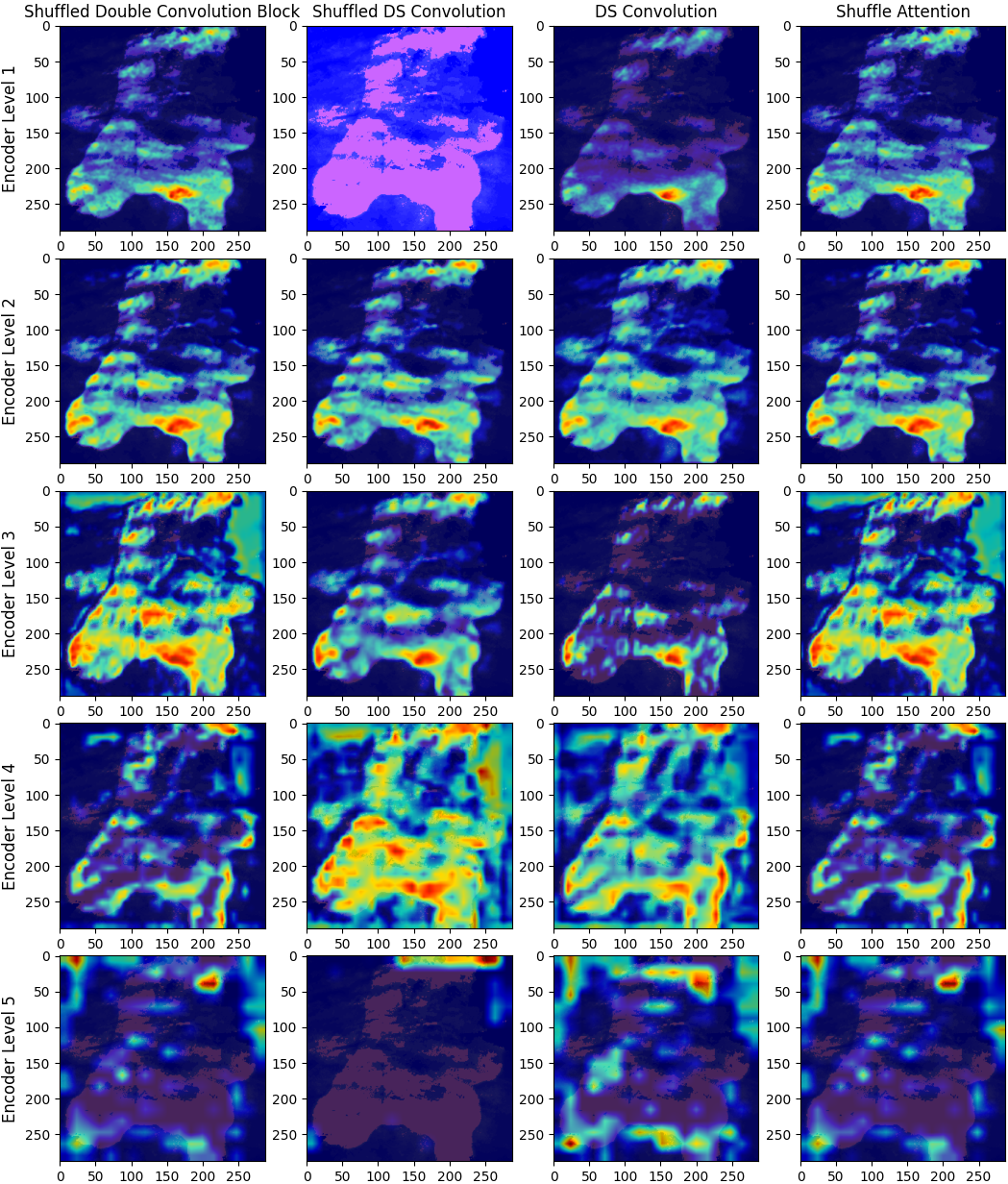}
    \caption{Heatmaps generated using Grad-CAM for the SSA-UNet model, showing activation regions across the five levels of the encoder. The visualizations include the heatmaps generated by the Shuffled the Double Convolution Block, the shuffled depthwise separable convolution layer, the depthwise separable convolution layer and the Shuffle Attention module.  }
    \label{gradcam2}
\end{figure}

\begin{figure}[h!]
    \centering
    \includegraphics[width=\linewidth]{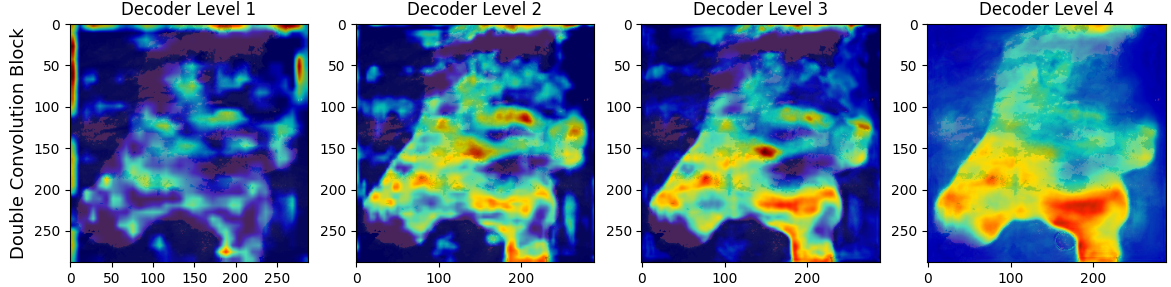}
    \caption{Heatmaps generated using Grad-CAM for the SSA-UNet model, showing activation regions across the four levels of the decoder. The visualizations include the heatmaps generated by the Double Convolution Block. }
    \label{gradcam}
\end{figure}

\subsection{Grad-CAM visualizations}

The analysis of heatmaps generated with Grad-CAM provides valuable insights into the model's predictive behavior. Fig. \ref{gradcam2} shows the heatmaps generated from the main component of the encoder branch using the same data as in Fig. \ref{fig:predictions}. Specifically, the behavior of the double convolution blocks, the shuffled and classic depthwise separable convolutions, and the Shuffle Attention module are examined. A similar pattern is observed across all layers as the levels deepen. At level 1, both the shuffled double convolution block and the shuffle attention module closely resemble the input map, while the depthwise separable convolution seems to focus only on the lower area of the map. At level 2, all modules generate similar heatmaps, indicating that they have a comparable impact on the final prediction. At level 3, the shuffled double convolution block and the Shuffle Attention module begin to focus on regions with little to no precipitation. Similarly, at level 4, the shuffled depthwise separable convolution and the depthwise separable convolution also start focusing on regions with low or no precipitation. This phenomenon has its peak at the last level, in which all the heatmaps generated are extremely scattered and focus on marginal areas. The same pattern can be identified on the heatmaps generated by the double convolution block of the decoder shown in Fig. \ref{gradcam}. As the depth increases, the double convolution block starts considering more marginal and scattered areas as more impactful for the final prediction. This tendency to place greater attention on more specific and scattered areas as the depth increases may result from the ability of deeper layers to capture more abstract and high-level features. Conversely, the shallower levels usually capture more low-level features, such as borders and shapes, making it reasonable for them to focus on the entire area of precipitation.

\section{Conclusion}

In this work, SSA-UNet is proposed, an architecture that implements channel shuffling within depthwise separable convolutions and an attention mechanism to enhance the model's overall performance. Thanks to the implementation of shuffled depthwise separable convolution and the Shuffle Attention mechanism, the number of trainable parameters is reduced while achieving equal or even superior predictive performance compared to SmaAt-UNet. Therefore, SSA-UNet results particularly suited for deployment on mobile devices or resource-constrained environments for weather forecasting applications. Furthermore, the explainability analysis using Grad-CAM heatmaps provided a partial elucidation of the model's prediction mechanism, offering valuable insights for understanding its outputs. It is worth mentioning that its compact size and fast inference time enable training on larger and more up-to-date datasets, even in limited computational settings. This not only makes its use more accessible, but also reduces the environmental footprint associated with model training, aligning with the growing demand for sustainable AI solutions.



\bibliographystyle{plain}
\bibliography{refs}

\end{document}